\documentclass{bmvc2k}

\title{OODformer: Out-Of-Distribution Detection Transformer}

\addauthor{Rajat Koner}{koner@dbs.ifi.lmu.com}{1}
\addauthor{Poulami Sinhamahapatra}{poulami.sinhamahapatra@iks.fraunhofer.de}{2}
\addauthor{Karsten Roscher}{karsten.roscher@iks.fraunhofer.de}{2}
\addauthor{Stephan G\"unnemann}{guennemann@in.tum.de}{3}
\addauthor{Volker Tresp}{volker.tresp@siemens.com}{14}

\addinstitution{
 Ludwig Maximilian University, Munich,\\ Germany
}
\addinstitution{
 Fraunhofer IKS, Fraunhofer Institute for Cognitive Systems IKS, Munich,\\ Germany
}
\addinstitution{
 Technical University of Munich,\\ Germany
}
\addinstitution{
 Siemens AG, Munich,\\ Germany
}

\runninghead{KONER ET AL}{OODformer: Out-Of-Distribution Detection Transformer}

\begin{document}

\maketitle
\sloppy

\begin{abstract}
A serious problem in image classification is that a trained model might perform well for input data that originates from the same distribution as the data available for model training, but performs much worse for 
out-of-distribution (OOD) samples. 
In real-world safety-critical applications, in particular, 
it is important to be aware if a new data point is OOD. 
To date, OOD detection is typically addressed using either confidence scores, auto-encoder based reconstruction, or contrastive learning. 
However, the global image context has not yet been explored to discriminate the non-local objectness between in-distribution and OOD samples.
This paper proposes a first-of-its-kind OOD detection architecture named OODformer that leverages the contextualization capabilities of the transformer. Incorporating the trans\-former as the principal feature extractor allows us to exploit the object concepts and their discriminatory attributes along with their co-occurrence via visual attention.
Based on contextualised embedding, we demonstrate OOD detection using both class-conditioned latent space similarity and a network confidence score. Our approach shows improved generalizability across various datasets. We have achieved a new state-of-the-art result on CIFAR-10/-100 and ImageNet30. Code is available at : \url{https://github.com/rajatkoner08/oodformer}.
\end{abstract}
\section{Introduction}
\label{sec:intro}
Deep learning has been shown to give excellent results when the data in an application comes from the same distribution as the data that was available for model training, also called in-distribution (ID) data. Unfortunately, performance might deteriorate drastically for out-of-distribution (OOD) data.  
The reason why application data might be OOD can be manifold and is often attributed to complex distributional shifts, the appearance of an entirely new concept, or random noise coming from a faulty sensor. 
As deep learning becomes the core of many safety-critical applications like autonomous driving, surveillance system, and medical applications,  distinguishing ID from OOD data is of paramount importance. 

The recent progress in generative modelling and contrastive learning has led to a sig\-nif\-i\-cant advancement in various OOD detection methods. \cite{liang2017enhancing,hsuGeneralizedODINDetecting2020a} improved \textit{softmax} distribution for outlier detection. \cite{tackCSINoveltyDetection2020,winkensContrastiveTrainingImproved2020b,sehwag2021ssd} used contrastive learning for OOD detection. The common idea in these works is that in a contrastively trained network, similar objects will have similar embeddings while dissimilar objects will be repelled by the contrastive loss. 
However, these approaches  often require fine-tuning with OOD data or rely on several negative samples and often suffer from inductive biases prevalent in convolution based  architectures. Hence, it is difficult to train and deploy them out-of-the-box. This motivates us to go beyond the conventional practice of using negative samples or inductive biases in designing an OOD detector. Along this line, we argue that systematic exploitation of global image context presents a potential alternative to obtain a semantically compact representation. 
To systematically exploit the global image context, we leverage the multi-hop context accumulation of the vision-transformer (ViT) \cite{dosovitskiy2020image}.


\begin{wrapfigure}{R}{7cm}
\centering
\includegraphics[width=0.55\textwidth,height=4cm]{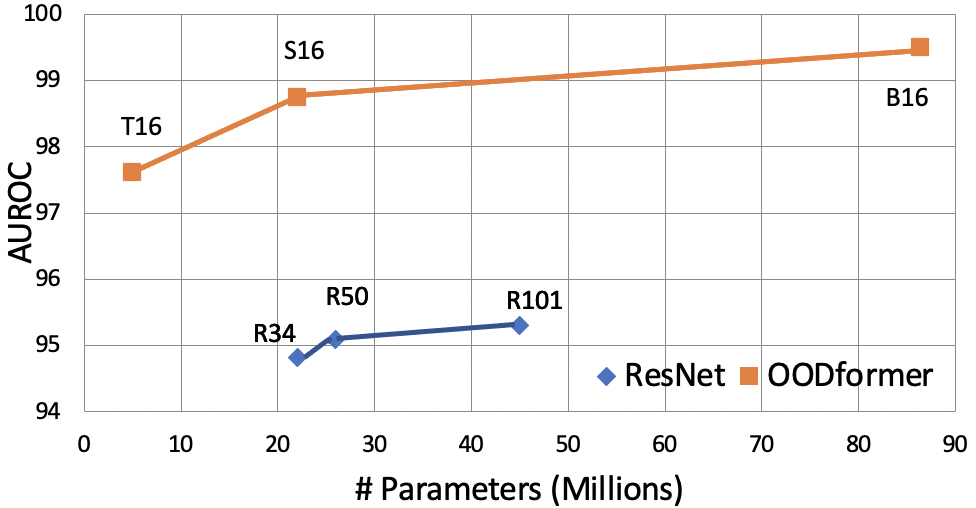}
\caption{Comparison of Transformer with traditional ResNet based variant, both trained with ID (CIFAR-10) samples, for distinguishing ID vs OOD (SVHN) samples .}
\label{fig:auroc_vs_param}
\end{wrapfigure}

Transformer's  visual attention sig\-nif\-i\-cantly outperforms convolutional architectures on various image-classification \cite{dosovitskiy2020image,touvron2020deit}, object-detection \cite{carion2020end}, relationship-de\-tect\-ion \cite{koner2020relation,koner2021scenes} and other vision-oriented task \cite{liu2021swin,hildebrandt2020scene,caron2021emerging,koner2021graphhopper}. However, transformer's ability to act as a generalized OOD detector remains unexplored so far. As a suitable transformer candidate, we investigate the emerging transfomer architecture in image-classification tasks, namely the  Vision Transformer (ViT) \cite{dosovitskiy2020image}, and its data-efficient variant DeiT \cite{touvron2020deit}.
ViT explores the global context of an image and its feature-wise correlations that are extracted from small image patches using visual attention. 
Intuitively, the difference between an ID and OOD is a small-to-large perturbation in form of the incorrect ordering or corruption (incl. insertion or deletion). Thus, we argue that capturing class/object attribute interaction is important for OOD detection, which we have implemented via a transformer.

To the best of our knowledge, we are the first to propose and investigate the role of global context and feature correlation using the vision transformer to generalise data in terms of OOD detection. Our key idea is to train ViT with an ID data set and use the similarity of its final representative embedding for outlier detection. 
Through the patch-wise attention on the attributes, we aim to reach a discriminatory embedding that is able to distinguish between ID and OOD. First, we follow a formal supervised training of ViT with an in-distribution data set using cross-entropy loss. In the second step, we extract the learned features from the classification head ($[class]$ token) and calculate a class conditioned distance metric for OOD detection. Our experiment exhibits that \textit{softmax} confidence prediction from the ViT is more generalisable in terms of OOD detection than the same from the convolutional counterpart. We also observe that the former overcomes  multiple shortcomings of the latter, e.g., poor margin probabilities \cite{elsayed2018large,liu2016large}. 
Figure \ref{fig:auroc_vs_param}, illustrates our claim that, even with much fewer parameters, transformer-based architecture performs significantly better compared to traditional convolution-based architecture. This shows the superiority of the transformer in the outlier detection task.

We evaluate the effectiveness of our model on several data sets such as CIFAR-10, CIFAR-100 \cite{krizhevsky2009learning}, ImageNet30 \cite{hendrycks2019using} with multiple setting (e.g., class conditioned similarity, \textit{softmax} score). Our model outperforms the convolution baseline and other state-of-the-art models in all settings by a large margin. Finally, we have conducted an extensive ablation study to strengthen the understanding of the generalizability of ViT and the impact of the shift in data.
In summary, the key contributions of this paper are:

\begin{itemize}

\item We model the OOD task as an object-attribute-based compact semantic representation learning. In this context, we are the first to propose a vision-transformer-based OOD detection frame\-work named OODformer and thoroughly investigate its efficiency.

\item We probe extensively into the learned representation to verify the proposed method's efficacy using both class-conditioned embedding-based and \textit{softmax} confidence score-based OOD detection. 

\item We provide an in-depth analysis of the learned attention map, softmax confidence, and embedding properties to intuitively interpret the OOD detection mechanism.

\item We have achieved state-of-the-art results for numerous challenging datasets like CIFAR-10, CIFAR-100, and ImageNet30 with a significantly large gain. 

\end{itemize}


\section{Related Work}
\label{sec:related}
OOD detection approaches can be classified into a number of  categories. The first and most intuitive approach is to classify the OOD sample using a   confidence score derived from the network.  \cite{hendrycksBaselineDetectingMisclassified2018a} propose max \textit{softmax} probability,  consecutively improved by ODIN using temperature scaling \cite{liang2017enhancing}. \cite{leeSimpleUnifiedFramework2018} utilised Mahalanobis distance and \cite{hsuGeneralizedODINDetecting2020a} improved ODIN without using OOD data  even further. Other density approximation-based generative model\-ling \cite{renLikelihoodRatiosOutofdistribution2019, serraInputComplexityOutofdistribution2019, nalisnickDeepGenerativeModels2019} also contributed in this approach. \cite{postnet} also shown density based uncertainty estimation could also be used for OOD detection.  A second direction towards OOD detection is to train a generative model for likelihood estimation. \cite{zong2018deep,pidhorskyiGenerativeProbabilisticNovelty2018} focuses on learning represent\-ations of training samples, using a  bottleneck layer for efficient reconstruction and generalis\-ation of samples. Another advancement on OOD detection is based on self-supervised \cite{chenSimpleFrameworkContrastive2020c,chenBigSelfSupervisedModels2020,liPrototypicalContrastiveLearning2020,chenImprovedBaselinesMomentum2020a} or supervised \cite{khoslaSupervisedContrastiveLearning2020,chuangDebiasedContrastiveLearning2020} contrastive learning. It aims on learning effective discriminatory represent\-ations by forcing the network to learn similar representations for similar  semantic classes while repelling the others. This property has been utilised by many recent works \cite{tackCSINoveltyDetection2020,winkensContrastiveTrainingImproved2020b,sehwag2021ssd} for OOD detection. The common idea in these works is that in a contrastively trained network, semantically closer objects from ID samples will have similar representation, whereas OOD samples would be far apart in the embedding space.
\section{Attention-based OOD detection}
\label{sec:methods}
In this section, the OOD detection problem is presented followed by a brief background about vision transformer, and the feature similarity-based outlier detection. As OOD labels may not be available for most scenarios, our method primarily relies on a similarity score-based detection. 

\subsection{Problem Decomposition : OOD Detection}
Let $x_{in} \in X \subseteq \mathbb{R}^k$ a  training sample
with $k^{th}$ dimensions and let  $y_{in} \in Y=\{1,...,C\}$ 
be its class label, where $C$ is number of classes. For a given neural feature extractor ($F_{feature}$), learned feature vector $X_{feat} \subseteq \mathbb{R}^d$ is obtained as $X_{feat} = F_{feature}(X)$. Finally we get posterior class probabilities as $P(y=c|x_{feat})$, where $x_{feat} \in X_{feat}$ and $y \in Y =F_{classifier}(X_{feat}) $. $F_{feature}$ and $F_{classifier}$ are two learned functions that map an image from data space to feature space and then derive the  posterior probability distribution. In a real world setting, data drawn from $X$ in an application may not follow the same distribution as the  training samples ($x_{in}$).  We refer these data as  OOD ($x_{ood} \in X,p(x_{ood}) \neq p(x_{in})$). 
A question now is  to what extent OOD data   diverges from an ID in their representations? A second question is the reliability of the prediction of the posterior probability distributions for OOD data. 
To quantify the shift in data from the ID samples, we compute the similarity of the embedding between the samples ($x_{in}$ or {$x_{out}$}) and the nearest ID class mean.  In an ideal scenario, this representational similarity should be much less for OOD ($x_{out}$) than ID ($x_{in}$). Also, its softmax confidence should be significantly lower than that of ID samples, allowing the use of a simple threshold to distinguish between ID and OOD samples.

\subsection{Feature extraction: Vision Transformer}

In our work we employ the Vision Transformer (ViT) \cite{dosovitskiy2020image} and its data efficient variant DeiT \cite{touvron2020deit}; 
they use an identical transformer encoder \cite{vaswani2017attention} architecture with the  same configura\-tion. However, ViT uses ImageNet-21K for pre-training, whereas DeiT uses more robust data augmentation and is only trained with ImageNet. The encoder of ViT takes only a  1D sequence of tokens as input. However, to handle a 2D image with height $H$ and width $W$, it divides the image into $N$ number of small 2D patches and flattened into a 1D sequence $X \in \mathbb{R}^{N\times (P^2 \cdot C')}$, where $C'$ is the number of channels, $(P, P)$ is the resolution of each patch and $N$ is the number of sequences obtained as $N=HW/P^2$. A  $[class]$ token ($x_{cls}$) similar to BERT \cite{devlin2018bert} prepended (first position in sequence) to the sequence of patch embeddings,  as expressed in Eq.\ref{eq:vit}, 
\begin{equation}
\label{eq:vit}
    z_{x} = [x_{cls};x_p^1E;x_p^2E;...;x_p^NE] + E_{pos}, E \in \mathbb{R}^{(P^2 \cdot C') \times d },E_{pos} \in \mathbb{R}^{(N+1) \times d }\\
\end{equation}
where $x_{cls},E_{pos}$ are learnable embedding of $d^{th}$ dimension. $E$ is the linear projection layer of $d$ dimension for a patch embedding (e.g., $x_p^1E$). The $[class]$ token embedding ($x_{feat} \in \mathbb{R}^{d}$) from the final layer of the encoder is used for classification. This classification token  serves as a representative feature from all patches accumulated using global attention.

At the heart of each encoder layer (for details, see supplementary) lies a multi-head self-attention (MSA) and a multi-layer perception (MLP) block. The MSA layers provide global attention to each image patch; thus, it has less inductive bias for local features or neighbourhood patterns than a CNN. One image patch or a combination thereof represents either a semantic or spatial attribute of an object. Therefore, we hypothesise that cumulative global attention on these patches would be helpful to encode discriminatory object features. Positional embedding ($E_{pos}$) of the encoder is also beneficial for learning the relative position of a feature or a patch. Subsequently, a local MLP block makes those image features translation invariant. This combination of MSA and MLP layers in an encoder jointly encodes the attributes' importance, associated correlation, and co-occurrence. In the end, $[class]$ token, being a representative of an image  $x$, consolidate multiple attributes and their related features via the global context, which is helpful to define and classify a specific object category.          

The [$class$] token from the final layer is used for OOD detection in two ways; first, it is passed to $F_{classifier}(x_{feat})$ for \textit{softmax} confidence score, and second it is used for latent space distance calculation as described in next section. 

\subsection{OOD detection} 
 For an OOD sample, one or more attributes of the object are assumed to be different (e.g. corrupted or new attributes). Hence, an OOD sample should lie in a distribution that has significant distributional shift from training samples. We assume these shifts in data would be captured by the representational features extracted from the ViT. We quantify this variation of features or attributes compared to ID samples using 1) a distance metric on the latent space embedding and 2) \textit{softmax} confidence score.

\begin{algorithm}[H]
\footnotesize{}
\SetAlgoLined
\textbf{Input:} training samples $\{x^i_{in}\}$ and training labels $\{y^i_{in}\}$ for $i=1:n$, test sample $x_{test}$\\
\textbf{Output:} $x_{test}$ is a outlier or not?\\
\For{each class: $c\gets1$ \KwTo $C$}{
 $\{x^i_{feat}\}_{i=1:n_c}=F_{feature}(\{x^i_{in}|y^i_{in}=c\}_{i=1:n_c})$, where $n_c=|\{x^i_{in}|y^i_{in}=c\}|$\\
    compute  mean :  $\mu_{c} = \frac{1}{n_c} \sum_{i=1}^{n_c} (x^i_{feat})$\\
    compute  covarience : $\Sigma_c = \frac{1}{n_c-1}\sum_{i=1}^{n_c}(x^i_{feat}-\mu_{c})(x^i_{feat}-\mu_{c})^\top$

    }
$x_{feat}^{test} = F_{feature}(x_{test})$\\
$distance =  \min_{c}( \sum_{c=1}^{C}(x_{feat}^{test} - \mu_{c})  \Sigma_{c}^{-1} (x_{feat}^{test} - \mu_{c})^\top)$ \\
$conf = \max_{c}(softmax(F_{classifier}(x_{feat}^{test}))$\\
\eIf {$(distance>t_{distance}) \ OR \ (conf<t_{conf)}$}{
$x_{test} $is an outlier
}{
$x_{test}$ is not an outlier
}
 \caption{OOD detection using distance metric}
 \label{algo:algo1}
\end{algorithm}

\noindent\textbf{Distance in Latent Space:} Multiple attributes of an object can be found in different spatial locations of an image. In vision transformer, image patches are ideal candidates for representi\-ng each of the individual attributes. Global information contextualisation from these attributes plays a crucial role in the classification of an object (see Sec. \ref{sec:abl}). Supervised learning (e.g.,cross-entropy) should benefit by an accumulation of object-specific semantic cues through $[class]$ token for such global attributes and their context. This incentivises implicit clustering of object classes that have similar attributes and features, which are favourable for generalisation. To take advantage of attribute similarities, we compute class-wise distance over the activation of $[class]$ token. First, we compute the mean ($\mu_{c}$) of all class categories present in the training samples. Second, for a test sample, we compute the distance between its embedding from the mean embedding of each class. Finally, the test sample is classified as OOD if its distance is more than the threshold ($t_{distance}$) to its nearest class. We have used the Mahalanobis distance metric for our experiment. The output from a representative $[class]$ token is normalised with the transformer default layer normalisation \cite{ba2016layer} for every token. It makes the embedding distribution normal, thus Mahalanobis distance could utilise the normally distributed mean and co-variance unlike Euclidean distance which only uses mean. Mahalanobis distance from a sample to distribution of mean $\mu_{c}$ and covariance $\Sigma_{c}$ can be defined as
\begin{equation}
    \label{eq:mahalanobis}
     D_c(x_{feat}^{test}) = (x_{feat}^{test} - \mu_{c})  \Sigma_{c}^{-1} (x_{feat}^{test} - \mu_{c})^\top  .
\end{equation}
We have also examined Cosine and Euclidean distances as shown in Figure \ref{fig:main_abl}a. The complete algorithm to compute an outlier has been given to Algorithm \ref{algo:algo1}.

\noindent\textbf{Softmax confidence score: } We obtain the final class probabilities from $F_{classifier}$ through \textit{softmax}. Softmax-based posterior probabilities have been reported in earlier studies to give erroneous high confidence score when exposed to outliers \cite{leeSimpleUnifiedFramework2018}.  Prior works used this posterior probability for OOD detection task, using either a binary classifier \cite{hendrycksBaselineDetectingMisclassified2018a} or temperature scaling \cite{liang2017enhancing}. We argue that for a good attribute cluster representation, as in the case of a transformer, no extra module for OOD detection is needed (see Table. \ref{tab:ResNet_comp}). Thus in this work, we use only simple numerical thresholds ($t_{conf}$) from ID samples for the detection of outliers without using additional OOD data for fine-tuning.

\section{Experiments}
\label{sec:exp}
\begin{table}[]
\scriptsize
\centering
\begin{tabular}{c|c|c|c|c|c|l}
\toprule
ID                       & \multicolumn{2}{c|}{CIAFR-10}                                   & \multicolumn{2}{c|}{CIFAR-100}                                   & \multicolumn{2}{c}{IM-30}                                      \\ \midrule
(Out-of-Distribution)    & SVHN                           & CIFAR-100                       & SVHN                           & CIFAR-10                         & CUB                            & Dogs                           \\ \midrule
Baseline OOD\cite{hendrycksBaselineDetectingMisclassified2018a}              & 95.9                           & 89.8                           & 78.9                           & 78.0                            & -                              & -                              \\
ODIN\cite{liang2017enhancing}                      & 96.4                           & 89.6                           & 60.9                           & 77.9                            & -                              & -                              \\ 
Mahalanobis\cite{leeSimpleUnifiedFramework2018}               & 99.4                           & 90.5                           & 94.5                           & 55.3                            & -                              & -                              \\ 
Residual Flows\cite{zisselman2020deep}            & 99.1                           & 89.4                           & 97.5                           & 77.1                            & -                              & -                              \\ 
Outlier exposure\cite{hendrycksDeepAnomalyDetection2019}          & 98.4                           & 93.3                           & 86.9                           & 75.7                            & -                              & -                              \\ 
Rotation pred\cite{hendrycksUsingSelfSupervisedLearning2019}             & 98.9                           & 90.9                           & -                              & -                               & -                              & -                              \\ 
Contrastive + Supervised\cite{winkensContrastiveTrainingImproved2020b}  & 99.5                           & 92.9                           & 95.6                           & 78.3                            & 86.3                           & 95.6                           \\ 
CSI\cite{tackCSINoveltyDetection2020}                        & 97.9                           & 92.2                           & -                              & -                               & 94.6                           & 98.3                           \\ 
SSD+\cite{sehwag2021ssd}                     & \textbf{99.9} & 93.4                           & 98.2                           & 78.3                            & -                              & -                              \\ \midrule
OODformer(Ours)          & 99.5                           & \textbf{98.6} & \textbf{98.3} & \textbf{96.1} & \textbf{99.7} & \textbf{99.9} \\ \bottomrule
\end{tabular}
\vspace{1em}
\caption{Comparison of OODformer with state-of-the-art detectors trained with supervised loss.}
\vspace{-2em}
\label{tab:main_comp}
\end{table}

In this section, we evaluate the performance of our approach and compare it to  state-of-the-art OOD detection methods. In Sec. \ref{sec:result}, we report our results on labeled multi-class OOD detection and one class anomaly detection.  It also contains a  comparison with state-of-the-art methods and a  ResNet \cite{he2016deep} baseline. In Sec.\ref{sec:abl}, we examine the influence of architectural variance and distance metric on the  OODformer in the context of the OOD detection.

\noindent\textbf{Area under PR and ROC :} A single threshold score may not scale across all the data sets. In order to homogenise the performance across multiple data sets, a range of thresholds should be considered. Thus, we report the area under precision-recall (PR) and ROC curve (AUROC) for both the latent space embedding, as well as \textit{softmax} score.

\noindent\textbf{Setup:} We use ViT-B-16 \cite{dosovitskiy2020image} as the base model for our all experiments and ablation studies. Variants of ViT (Base-16, Large-16 with embedding size 768 and 1024 along with patch size 16) and DeiT (Tiny-16, Small-16 with corresponding embedding size of 192 and 384) used a similar architecture but differ on embedding and the number of attention heads (full details provided in appendix). As transformers have inherently lesser inductive biases compared to CNNs, they can exploit similarities more efficiently when pre-trained with large datasets as prevalent with well-known architecture like BERT \cite{devlin2018bert} or GPT2 \cite{radford2019language}. Thus, pre-training with large datasets is a necessary precondition for most transformer-based architectures and their use cases.
Hence, all the  models we use are pre-trained on ImageNet \cite{5206848} as recommended in \cite{dosovitskiy2020image}. Supervised cross-entropy loss is used for training along SGD as an optimizer and we followed the same data augmentation strategy like \cite{khoslaSupervisedContrastiveLearning2020}. ResNet-50  \cite{he2016deep} is our default convolution-based baseline architecture used across all result and ablation studies. Mahalanobis distance-based score on representational embedding is used as the default score for the AUROC calculation.\\
\noindent\textbf{Datasets:} We trained our networks on the following in-distribution data sets: CIFAR-10/-100 \cite{krizhevsky2009learning} and ImageNet-30 (IM-30) \cite{hendrycks2019using}. AS OOD data for CIFAR-10/-100 we have chosen resized Imagenet (ImageNet\_r) \cite{liang2017enhancing} and LSUN \cite{hendrycks2019using}  and SVHN \cite{netzer2011reading} as specified in \cite{tackCSINoveltyDetection2020}. In case of IM-30 we flow the same setup of \cite{tackCSINoveltyDetection2020}, and used Places-365 \cite{zhou2017places}, DTD \cite{cimpoi2014describing}, Dogs \cite{khosla2011novel}, Food-101 \cite{bossard2014food}, Caltech-256 \cite{griffin2007caltech}, and CUB-200 \cite{wah2011caltech}. Details about these data set are given in the appendix.
\linespread{0.5}
\subsection{Results}
\label{sec:result}
\textbf{Comparison with State-of-the-Art:}  Table \ref{tab:main_comp}, exhibits that even with a simple cross-entropy loss, OODformers achieve new state-of-the-art results on all data sets. Most importantly, OODformer supersedes its predecessor on the complex near OOD data sets, which strongly affirms substantial improvements in generalisation. In particular, detection of CIFAR-100 as OOD samples with a network trained in CIFAR-10 (ID) or vice-versa is the most challenging task since they share a significant amount of common attributes in similar classes. For example, a `truck' is an ID sample in CIFAR-10, whereas a `pickup-truck'  from CIFAR-100 is an OOD sample, despite being semantically closer and having many similar attributes. These attributes vs class trade-off leads to a significant drop in AUROC value for previous methods, especially when CIFAR-100 is ID and CIFAR-10 is OOD. We gained a notable \textbf{$5.2\%$} and \textbf{$17.8\%$} gain in AUROC when trained on CIFAR-10 and CIFAR-100, respectively. This can also be visualised in Figure.\ref{fig:main_abl}b. These results show that the OODformer has superior generalizability of the representational embedding due to the global contextualization of object attributes from single or multiple patches, even with complex near OOD classification.

\noindent\textbf{One class anomaly detection:} Similar to OOD detection, anomaly detection is concerned with certain data types or specifically one class. Apart from that, the specific class presence of any class will be considered an anomaly. In this setting, we consider one of the classes from CIFAR-10 as an in-distribution, and the rest of the classes are anomalies or outliers. We train our network with one class (ID) and the remaining nine classes become OOD and we repeat these experiments for every class, similar to \cite{tackCSINoveltyDetection2020}.  Table \ref{tab:outlier_comp},  shows our OODformer outperforms all existing methods and achieves new state-of-the-art results.
\begin{table}[t]
\scriptsize
\centering
\begin{tabular}{l|l|l|l|l|l|l|l|l|l|l|l}
\toprule
Methods   & Airplane & Automobile & Bird & Cat  & Dear & Dog  & Frog & Horse & Ship & Truck & Average\\ \midrule
GT\cite{golanDeepAnomalyDetection2018}        & 74.7     & 95.7       & 78.1 & 72.4 & 87.8 & 87.8 & 83.4 & 95.5  & 93.3 & 91.3 & 86.0 \\ 
Inv-AE\cite{huang2019inverse}    & 78.5     & 89.8       & 86.1 & 77.4 & 90.5 & 84.5 & 89.2 & 92.9  & 92   & 85.5 & 86.6 \\ 
Goad\cite{bergman2020classification}      & 77.2     & 96.7       & 83.3 & 77.7 & 87.8 & 87.8 & 90   & 96.1  & 93.8 & 92 & 88.2   \\ 
CSI\cite{tackCSINoveltyDetection2020}       & 89.9     & \textbf{99.9}       & 93.1 & 86.4 & 93.9 & \textbf{93.2} & 95.1 & 98.7  & 97.9 & 95.5  & 94.3\\ 
SSD\cite{sehwag2021ssd}       & 82.7     & 98.5       & 84.2 & 84.5 & 84.8 & 90.9 & 91.7 & 95.2  & 92.9 & 94.4  &90.0\\ \midrule
OODformer & \textbf{92.3}     & 99.4       & \textbf{95.6} & \textbf{93.1} & \textbf{94.1} & 92.9 & \textbf{96.2} & \textbf{99.1}  & \textbf{98.6} & \textbf{95.8} & \textbf{95.7} \\ \bottomrule
\end{tabular}
\vspace{1em}
\caption{Comparison of OODformer with other methods on one-class OOD from CIFAR-10.}
\label{tab:outlier_comp}
\vspace{-1.5em}
\end{table}

\begin{table}[]
\scriptsize
\centering
\begin{tabular}{c|c|c|c|c|c}
\toprule
\multirow{2}{*}{ID} &         \multirow{2}{*}{OOD}              & \multicolumn{2}{c|}{Emb-Distance} & \multicolumn{2}{c}{Softmax} \\
\cline{3-6}
        &  & ResNet        & OODformer        & ResNet      & OODformer     \\
\midrule 
\multirow{3}{*}{CIFAR-10}   & CIFAR-100                                     & 87.8          & \textbf{98.6}                            & 87.4        & \textbf{97.7}                         \\
                            & Imagenet\_r                                      & 91.4          & \textbf{98.8}                            & 90.9        & \textbf{96.0}                         \\
                            & LSUN                                          & 93.4          & \textbf{99.2}                            & 92.4        & \textbf{97.6}                         \\
\midrule
\multirow{3}{*}{CIFAR-100} & CIFAR-10              & 73.7          & \textbf{96.1}                            & 69.3        & \textbf{88.9}                         \\
                            & Imagenet\_r                                      & 79.9          & \textbf{92.5}                            & 72.4        & \textbf{86.1}                         \\
                            & LSUN                                          & 79.0          & \textbf{94.6}                            & 72.9        & \textbf{86.2}                         \\
\midrule
\multirow{6}{*}{IM-30}     & Places-365                                    & 82.9          & \textbf{99.2}                            & 91.8        & \textbf{98.2}                         \\
                            & DTD                                           & 97.8          & \textbf{99.3}                            & 90.9        & \textbf{98.2}                         \\
                            & Dogs                                          & 75.29         & \textbf{99.9}                            & 92.3        & \textbf{99.0}                         \\
                            & Food-101                                      & 73.44         & \textbf{99.2}                            & 83.2        & \textbf{97.2}                         \\
                            & Caltech256                                    & 86.37         & \textbf{98.0}                            & 91.4        & \textbf{96.8}                         \\
                            & CUB-200                                       & 87.22         & \textbf{99.7}                            & 91.9        & \textbf{99.4}                        \\
           \bottomrule
\end{tabular}
\vspace{1em}
\caption{Comparison of OODformer with ResNet baseline}
\label{tab:ResNet_comp}
\vspace{-2em}
\end{table}

\noindent\textbf{Comparison with ResNet: } Table \ref{tab:ResNet_comp}, shows the performance of OODformer in comparison with ResNet baseline. To show the generalisability and scalability, we trained our network with three in-distribution datasets and tested on nine different OOD datasets. Here, we use \textit{softmax} confidence score in addition to our default embedding distance-based score for OOD detection. As discussed in Sec \ref{sec:intro}, \textit{softmax} suffers from poor decision margin and lack of generalisation when used for ODD detection with CNNs. This can be addressed using our proposed OODformer. Table \ref{tab:ResNet_comp}, shows that ours \textit{softmax} based OOD detection significantly and consist\-ently outperforms ResNet and achieves an improvement as high as $19.6\%$. Our default AUROC score, using embedding similarity, also outperforms our baseline by a large margin. This result proves the objectness properties and exploitation of its related attributes through global attention play the most crucial role in  outlier detection. It bolsters our hypothesis that, for outlier detection, a transformer can serve as the de-facto feature extractor without any bells and whistles.

\begin{figure}[h]
\centering
\includegraphics[width=0.7\textwidth,height=4cm]{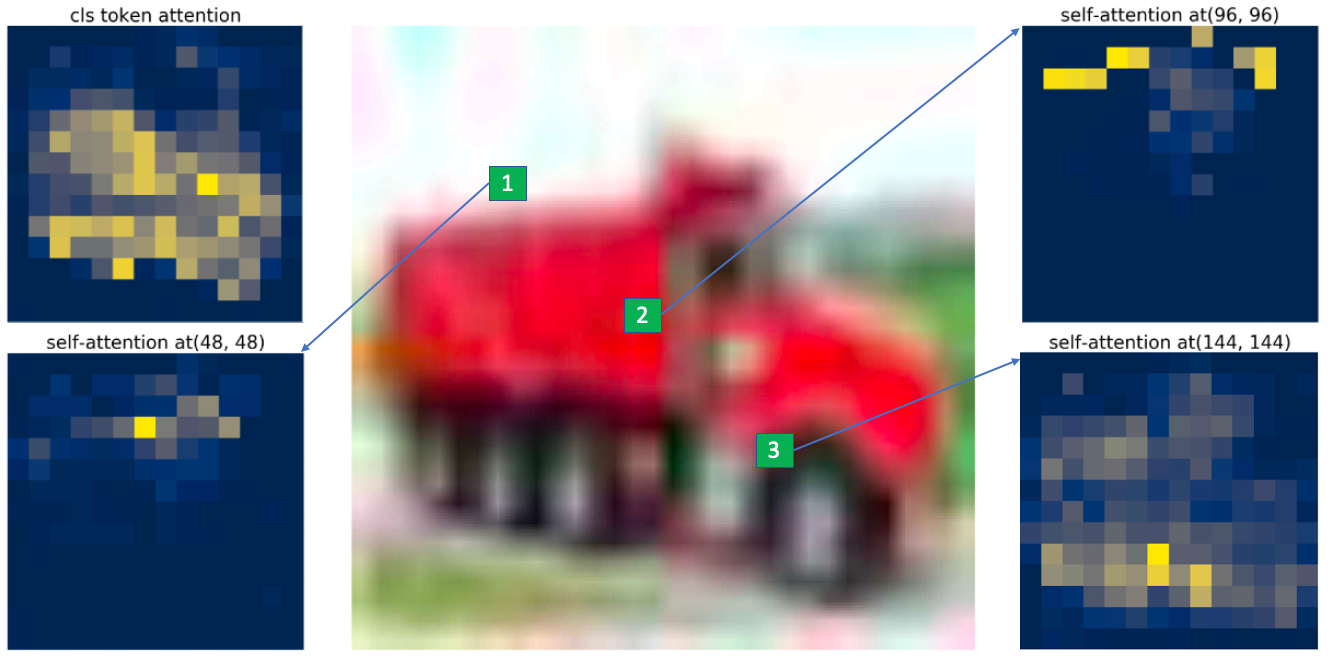}
\label{fig:subfig1}
\qquad
\includegraphics[width=0.15\textwidth]{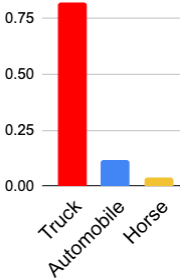}
\label{fig:subfig2}
\includegraphics[width=0.7\textwidth, height=4cm]{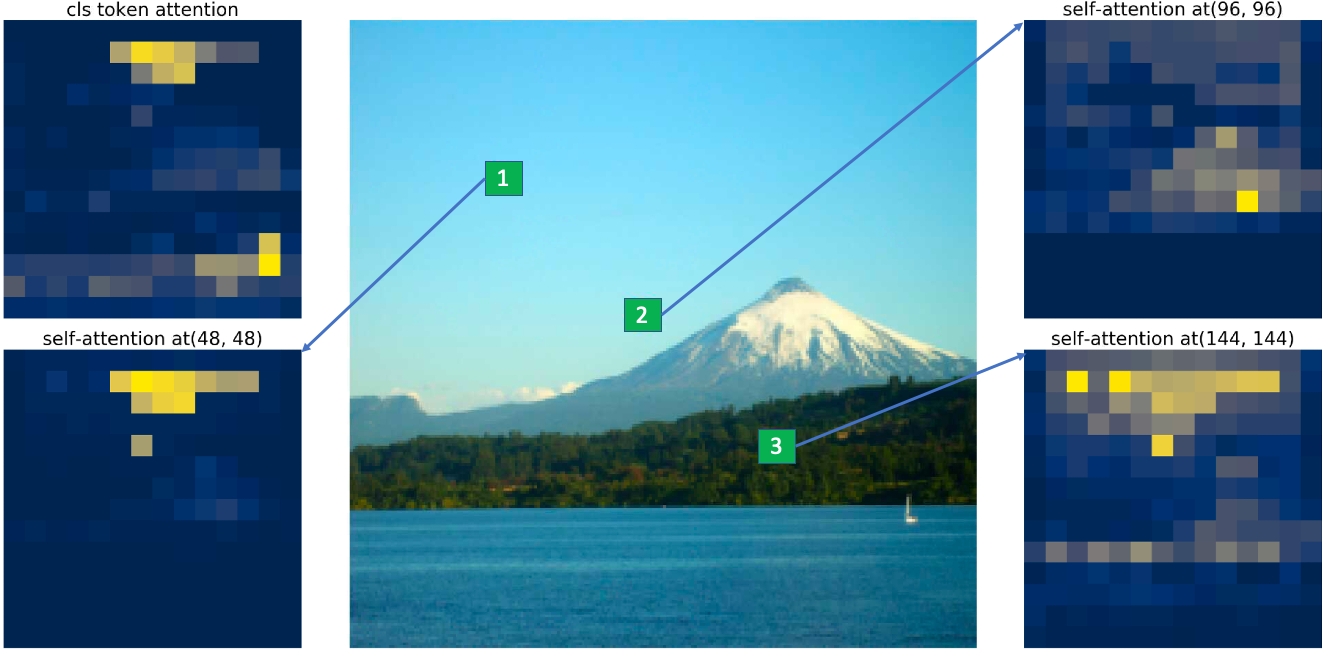}
\label{fig:subfig3}
\qquad
\includegraphics[width=0.15\textwidth]{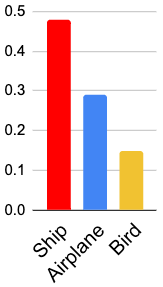}
\label{fig:subfig4}
\caption{Representative example of an ID sample of Truck (above), and OOD sample of Mountain (below). Among the four attention maps of each image, top left represents the class token, the remaining correspond to the image position marked with green square block. The right bar plot shows the top 3 similar class embedding.  }
\label{fig:globfig}
\vspace{-1.5em}
\end{figure}

\subsection{Dissection of OODformer}
\label{sec:abl}
In this section, we extensively analyse the attention and impact of different settings via an ablation study. We have used CIFAR-10 as ID and ViT-B-16 as default model architecture for our experiments.\\
\textbf{Global Context and Self-Attention:} To understand how vision transformers discriminate between an ID and OOD, we analyse its self-attention and the embedding space. Figure \ref{fig:globfig} depicts examples of both ID and OOD samples with their attention maps collected from the last layer. Self-attention accumulates information from all the related patches that define an object. For an ID sample (`truck'), $[class]$ token attention focuses mostly on the object of interest, while other selected patches put their attention based on properties similar to them like colour or texture. However, in the case of an OOD sample, the object `mountain' is unknown to the network, the $[class]$ token attention mostly focused on the sky and water, while we draw similar observations for other patches as before. This misplaced attention and absence of known object attributes leads to a lower similarity score and predicts wrong classes just because of background similarity. 
\begin{wraptable}{r}{7cm}
\scriptsize
\centering
\vspace{-0.5em}
\begin{tabular}{l|l|l|l|l}
\toprule
Model    & Acc. & CIFAR100 & Imagenet\_r & LSUN \\
\midrule
Resnet-34      & 95.6     & 87.2     & 89.7     & 91.4 \\
Resnet-50      & 95.4    & 87.5     & 90.0       & 91.7 \\
Resnet-101     & 95.7     & 87.8     & 91.4     & 93.4 \\
\midrule
Resnet-34(PT)     & 97.0     & 89.5     & 93.8     & 96.0 \\
Resnet-50(PT)      & 97.1    & 89.9     & 94.1       & 96.1 \\
Resnet-101(PT)     & 98.0     & 90.1     & 94.6     & 96.0 \\
\midrule
DeiT-T-16      & 95.4     & 94.4       & 95.2    & 97.3 \\
DeiT-S-16      & 97.6     & 96.6     & 96.3     & 98.4   \\
ViT-B-16      & 98.6     & 98.6     & 98.8     & 99.2 \\
ViT-L-16      & 98.8     & 99.1        & 99.2        & 99.4 \\ 
\bottomrule\\
\end{tabular}
\vspace{0.5em}
\caption{Comparison of various ResNet baseline (both not pre-trained and pre-trained) and vision transformer architecture. Here, (PT) defines the ResNet has been pre-trained on large scale ImageNet dataset.}
\vspace{-1em}
\label{tab:vitvsres}
\end{wraptable} 

\begin{figure}[ht]
\centering
\includegraphics[width=0.45\textwidth]{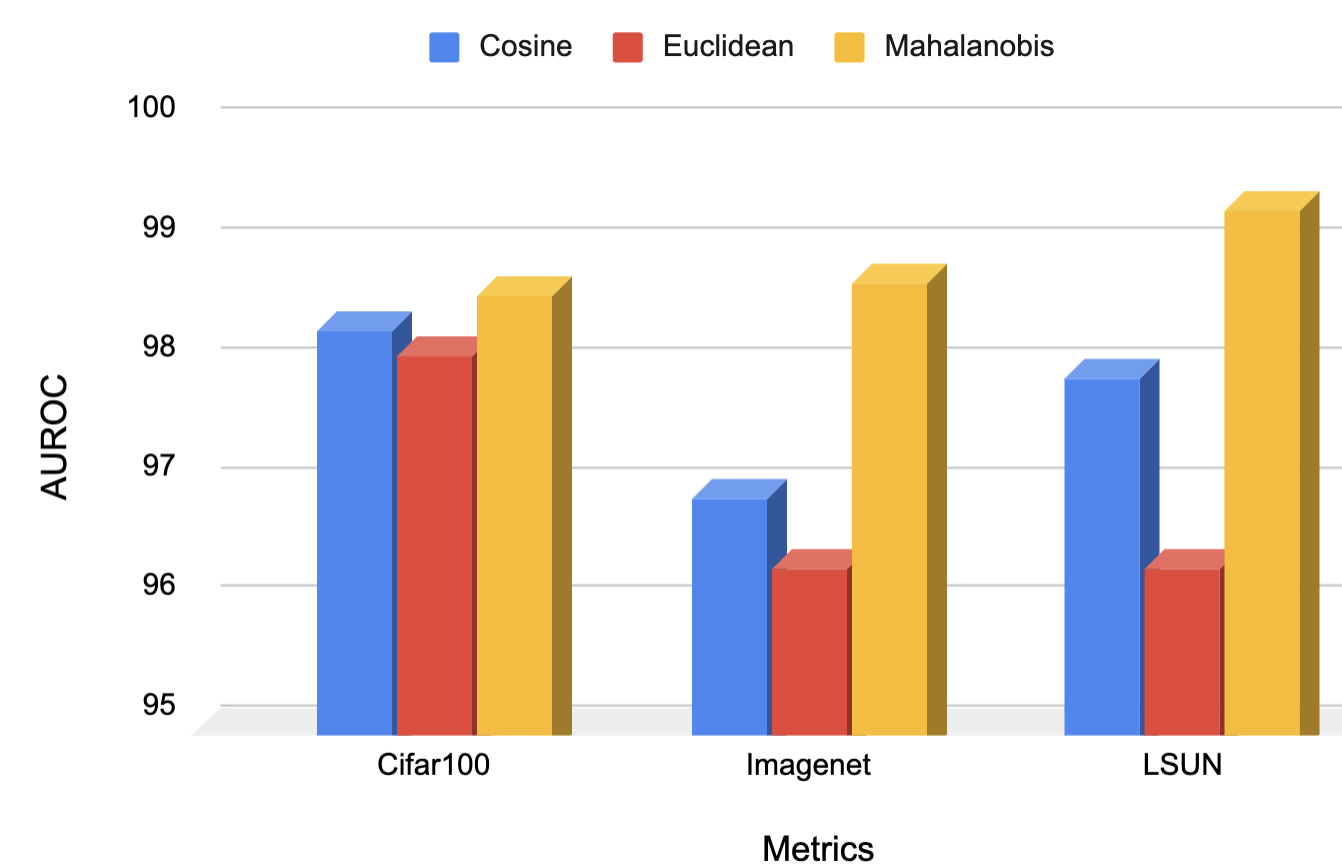}
\includegraphics[width=0.45\textwidth]{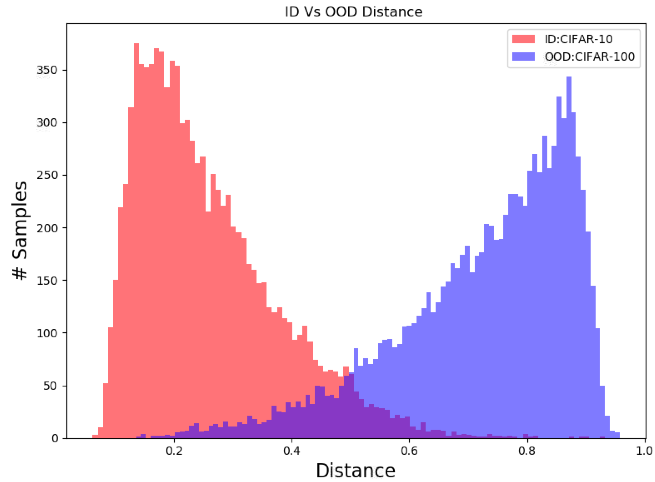}\\
\hspace{0.1cm}(a) \hspace{5cm} (b) 
\caption{Ablation Experiment : a) with distance metric, b) distance of ID and OOD samples from its corresponding class mean. }
\label{fig:main_abl}
\vspace{-1em}
\end{figure}

\noindent Figure \ref{fig:globfig}, also shows objectness and its related attributes exploration in a hierarchical way is crucial for OOD distance score. This can also be inferred from
Figure~\ref{fig:main_abl}b, where we notice that the transformer not only reduces intra-class distance for ID samples, it also increases the distance of OOD samples from the ID class mean.

\noindent\textbf{ResNets vs ViTs:} This section examines the effect of model complexity or expressiveness of a model for OOD or outlier detection. We have conducted experiments on multiple variants of ViT in comparison with variants of ResNet baseline (top rows) as shown in Table \ref{tab:vitvsres}. As all the variants of ViT and DeiT are pre-trained on ImageNet, for the sake of the comparison, we also evaluate ResNet variants that are pre-trained on ImageNet. As all the pre-trained ResNet models are trained with larger image size (224x224 pixel for ImageNet), thus for efficient utilisation of pre-training an up-scaling is needed. We fine-tune pre-trained ResNet on training in-distribution dataset using an up-scaling of 7x (from 32 to 224).
\begin{itemize}
    \item \textbf{ResNet Baseline :} Table \ref{tab:vitvsres}, shows that Deit-T-16, which is the lightest among all variants of ViT and much smaller in size than all ResNet baseline variants, performs substantially better on OOD detection(94.4 compared to 87.8 of R-101 for CIFAR-100). Furthermore, despite being similar in accuracy DeiT-T-16 is significantly superior with respect to OOD detection like all other ViT variants. 
    \item \textbf{ResNet Baseline pre-trained on ImageNet :} The transformer based architecture relies on training with large-scale data in order to achieve superior performance. ViT also requires large-scale pre-training on ImageNet for an efficient performance similar to a few well-known transformer based architectures like BERT\cite{devlin2018bert}, GPT2\cite{radford2019language}. More\-over, a few recent work \cite{reiss2021panda}, suggests pre-training of convolution offer better performa\-nce for outlier detection. Thus, to demonstrate the true performance of OODformer, we compare it with pre-trained convolutional models like ResNets (middle rows). Table \ref{tab:vitvsres}, suggests fine-tuning with upscale images offers good accuracy, but limited gain over OOD detection. Pre-trained ResNet follows a similar trend as the baseline models like increasing model capacity has a marginal effect on OOD detection in contrast to ViT variants. One of the key observations we found, apart from pre-training, up-scaling of images helped to attain higher performance on accuracy and OOD detection score.
\end{itemize}
 As per the above discussion based on Table \ref{tab:vitvsres}, we can clearly observe irrespective of their accuracy, image size, pre-training features extracted from ResNet deliver sub-optimal performance compare to all variants of ViT. Furthermore, with an increase in model complexity, the performance of ResNet reaches a plateau, whereas the performance of ViTs consistently improves with an increase in expressive power. This distinctly demonstrates the importance of objectness context, the role of an object attribute, and their mutual correlation (e.g, spatial) as hypothesised in the Sec \ref{sec:intro}.
Such observations contribute to our belief, that the transformer architecture is better suited for OOD detection tasks than classical CNNs primarily for less inductive bias, object attribute accumulation using attention.

\noindent\textbf{Analyzing Distance Metrics:}  We examine the influence of various distance metrics like Cosine and Euclidean in Figure.\ref{fig:main_abl}a. Euclidean utilises only the mean of the ID distribution thus it could be an intuitive reason for its underperformance. Furthermore, \cite{sehwag2021ssd} shows that, in embedding space, Euclidean distance is dominated by higher eigenvalues that reduce the sensitivity towards outliers. Both Cosine and Mahalanobis perform very similarly. However, the slightly better performance of Mahalanobis, compared to Cosine, could be attributed to its less dependency on a higher norm between two features and its utilisation of both mean and covariance. Figure \ref{fig:main_abl}a, shows that the OODformer is quite robust ($\pm 2\%$ variance) to various distances. 



\section{Conclusion}
\label{sec:conc}
\vspace{-1em}
In this paper, we made an early attempt utilising a vision transformer namely the OODformer for OOD detection. Unlike prior approaches, which rely upon custom loss or negative sample mining, we alternatively formulate OOD detection as object-specific attribute accumulation problem using multi-hop context aggregation by a transformer. This simple approach is not only scalable and robust but also outperforms all existing methods by a large margin. OODformer is also more suitable for being deployed in the wild for safety-critical applications due to its simplicity and increased interpretability compare to other methods. Building on our work, emerging methods in self-supervision, pre-training, and contrastive learning will be of future interest to investigate in combination with OODformer.

\section*{Acknowledgement}
We are sincerely thankful to Suprosnna Shit for his suggestion and revision of the manuscript.

\bibliography{egbib}
\newpage
\appendix
\section{ViT Architecture}
Vision Transformer \cite{dosovitskiy2020image} uses transformer encoder \cite{vaswani2017attention} for patch based image classification. The core of ViT relies on multi-head self-attention (MSA) and multi-layer perception (MLP) for processing sequence of image patches.
\paragraph{Multi-head Self-Attention:} The attention mechanism is formulated as a trainable weighted sum based approach. One can define self-attention as 
\begin{equation}
    \textit{Attention}(Q,K,V) = \textit{softmax}(\frac{QK^T}{\sqrt{d_k}})V
\end{equation}
where $Q,K,V$ are a set of learnable query, key and value and $d$ is the embedding dimension. A query vector $q \in \mathbb{R}^d$ is multiplied with key $r \in \mathbb{R}^d$ using inner product obtained  from the  sequence of tokens as specified in Eq. \ref{eq:vit}. The important features from the query token is dynamically learned by taking a \textit{softmax} on the product of query and key vectors. It is then multiplied with the value vector $v$ that incorporates features from other tokens based on their learned importance.
\paragraph{Multi-Layer Perception:} The transformer encoder uses a Feed-Forward Network (FFN) on top of each MSA layer. An FFN layer consists with two linear layer separated with GleU activation. The FFN processes the feature from the MSA block with a residual connection and normalizes with layer normalization \cite{ba2016layer}. Each of the FFN layer is local for every patch unlike the MSA (MSA act as a global layer), hence the FFN makes the encoder image translation invariant.
\section{Implementation Details}
Our backbone ViT \cite{dosovitskiy2020image} and DeiT \cite{touvron2020deit} are pretrained on ImageNet, and fine-tuned in an in-distribution dataset with SGD optimizer, a batch size of 256 and image size of  $224 \times 224$. We use a learning rate of $0.01$ with Cyclic learning rate scheduler \cite{smith2017cyclical}, weight decay=$0.0005$ and train for 50 epochs. We follow the data augmentation scheme same as \cite{khoslaSupervisedContrastiveLearning2020}. 
\subsection{Model Detail}
We use multiple variants of ViT and DeiT, primarily because DeiT offers lighter model, whereas ViT mainly focusses on havier model. The idea being an enhanced outlier detection performance with a lighter variant will bolster our assumption that exploring an object’s attributes and their correlation using global attention plays a crucial role in OOD detection. In comparison, a heavier variant will offer increased model capacity to improve the performance of the OODformer. Table. \ref{tab:vitvsres} exhibits the performance of OODformer with multiple backbone variants in support of our hypothesis. Specially the significant performance gain with the smallest variant of DeiT (T-16) bolster our claim. Table \ref{tab:deit_arch} shows the variation of their parameter, number of layers, hidden or embedding size, MLP size, number of attention head. 
\begin{table}[ht]
\centering
\begin{tabular}{c|c|c|c|c|c} 
\toprule
Model     & Prams & \#Layers & Hidden Size & MLP Size & \#Heads  \\ 
\midrule
DeiT-T-16 & 5     & 12       & 192         & 768      & 3        \\ 

DeiT-S-16 & 22    & 12       & 384         & 1536     & 6        \\ 

ViT-B-16  & 86.5  & 12       & 768         & 3072     & 12       \\ 

ViT-L-16  & 307   & 24       & 1024        & 4096     & 16       \\
\bottomrule\\
\end{tabular}\\
\caption{DeiT and ViT model architecture.}
\label{tab:deit_arch}
\end{table}
\subsection{Dataset Details}
Among the in-distribution dataset, 
CIFAR-10/-100 \cite{krizhevsky2009learning} consists of 50K training and 10K test images with corresponding 10 and 100 classes. The CIFAR-100 dataset also contains twenty superclasses for all the hundred classes present in it. Even though CIFAR-10 and CIFAR-100 has no overlap for any class, some classes share similar attributes or concepts (e.g., `truck' and `pickup-truck') as discussed in Section.\ref{sec:abl}. As a result of this close semantic similarity these two datasets poses the most challenging near OOD problem and the performance of OODformer in this context has shown in Table \ref{tab:main_comp}. Another in-distribution dataset, ImageNet-30 \cite{hendrycks2019using}, is a subset of ImageNet\cite{5206848} with 30 classes that contains 39K training and 3K test images.\\
Out-Of-Distribution dataset used for CIFAR-10/-100 are as follows : Street View Housing Number or SHVN \cite{netzer2011reading} contains around 26K test images of ten digits, LSUN \cite{hendrycks2019using} consists of 10K test images of ten various scenes, ImageNet-resize \cite{hendrycks2019using} is also a subset of ImageNet with 10K images and two hundred classes. For multi-class ImageNet-30, we follow the same OOD datsets as specified in \cite{tackCSINoveltyDetection2020}, they are : Places-365 \cite{zhou2017places}, Describable Texture Dataset \cite{cimpoi2014describing}, Food-101 \cite{bossard2014food}, Caltech-256 \cite{griffin2007caltech} and CUB-200 \cite{wah2011caltech}. 
\section{Ablation and Interpretation}
In addition to the analysis provided in Sec. \ref{sec:abl}, we ablate OODformer on various batch sizes, epochs and analyze the cluster in embedding space.\\
\begin{figure}[ht]
\centering
\includegraphics[width=0.45\textwidth]{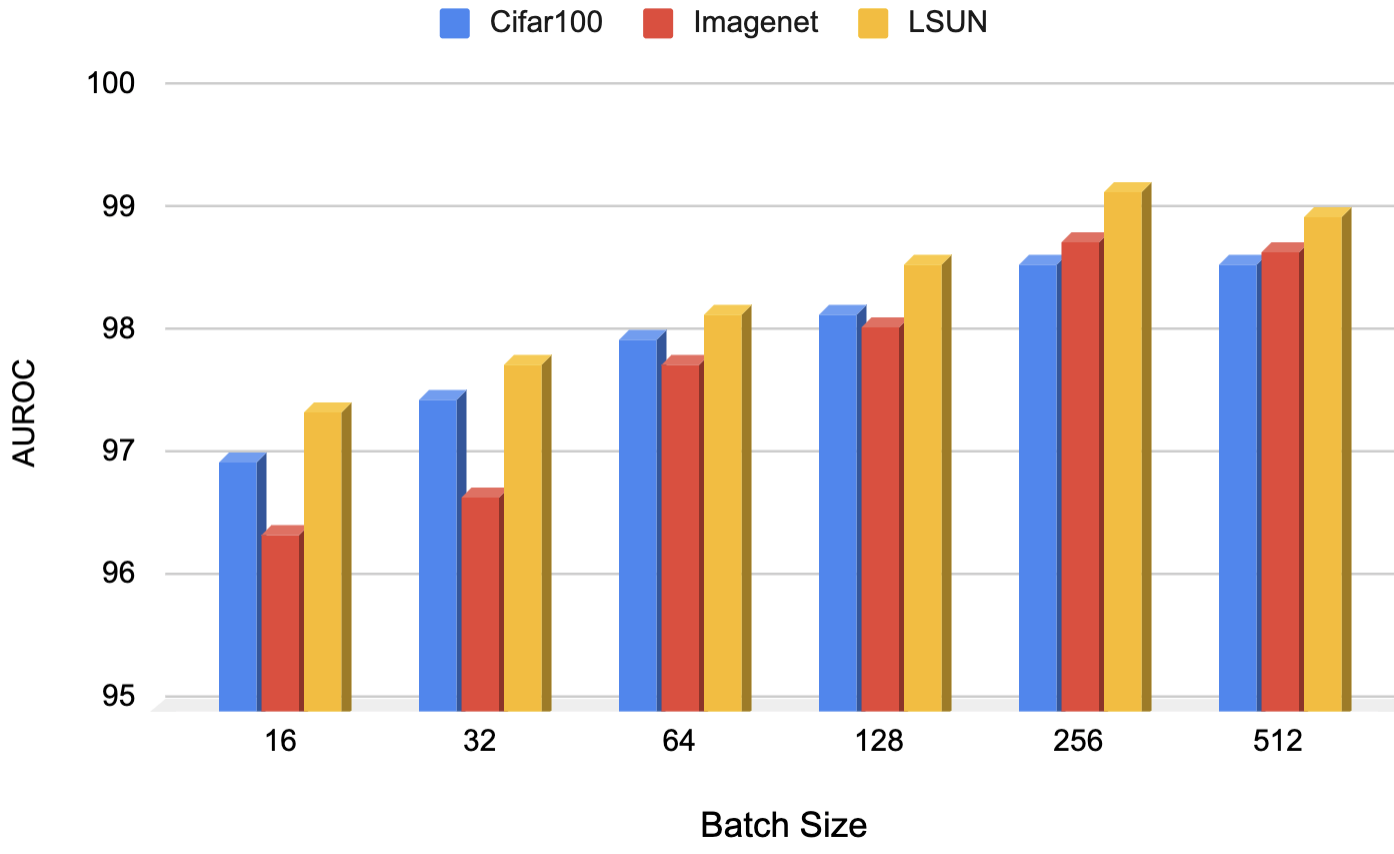}
\includegraphics[width=0.45\textwidth]{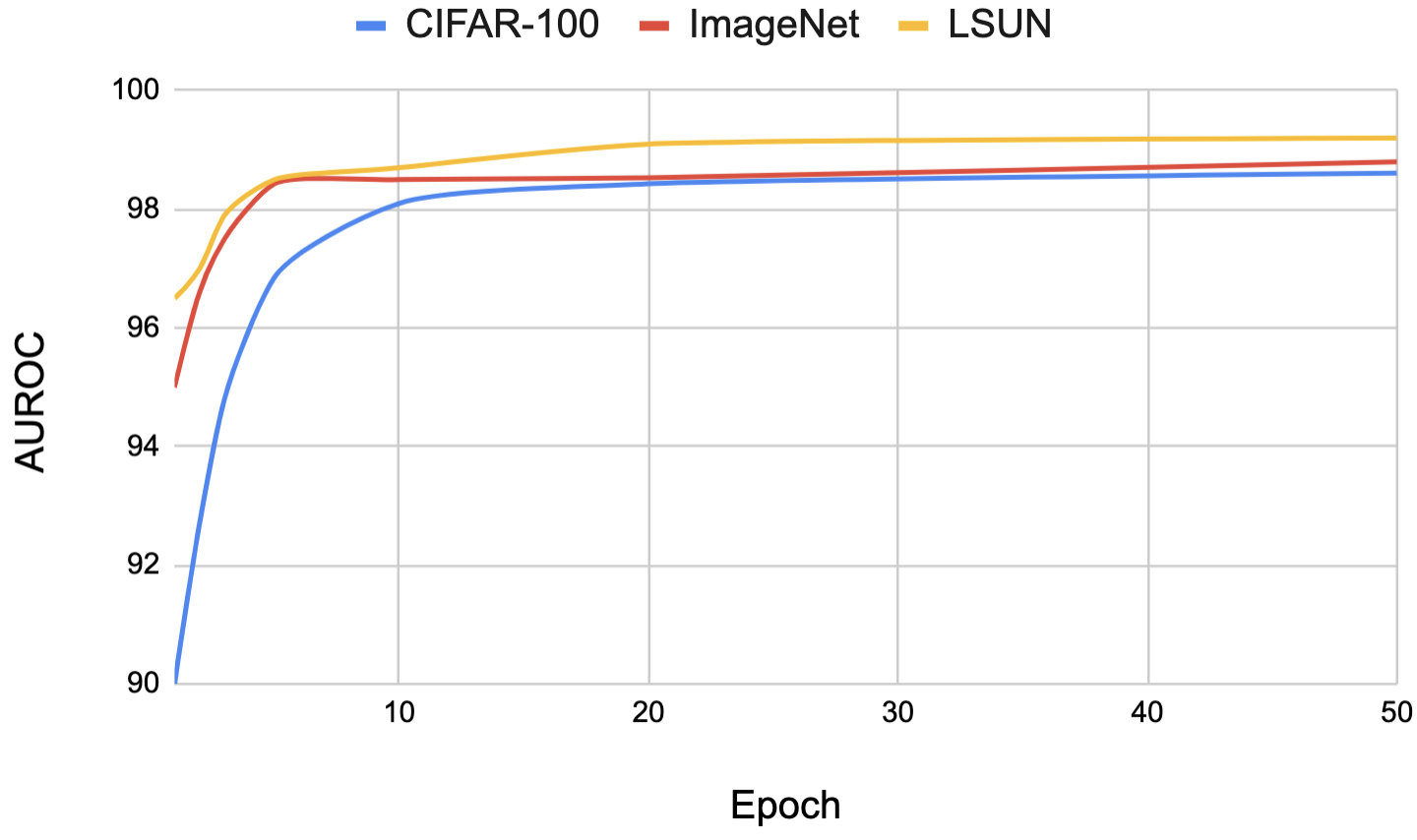}\\
\hspace{0.1cm}(a) \hspace{5cm} (b) 
\caption{Ablation Experiment : a) with various batch size, b) improvement of AUROC over the epochs. }
\label{fig:batch_epoch}
\end{figure}
Figure. \ref{fig:batch_epoch}a, demonstrates large batch size helps in OOD detection, though we observe it doesn't significantly impact accuracy on the in-distribution test set. An intuitive reason could be large batch size improves generalization \cite{hoffer2017train}, which enables the network to generalize object-specific properties that are helpful for outlier identification. Despite this gain, we observe OODformer remain relatively stable across all the batch sizes with OOD detection accuracy $\pm1.5\%$. However, the gain in AUROC gradually becomes stagnant with an increase of batch size suggest further scope of tuning learning rate is required using a linear scaling \cite{goyal2017accurate}. \\
Figure. \ref{fig:batch_epoch}b, shows an increase of outlier detection accuracy with the number of epochs. One of the important observation is easier OOD dataset (e.g., LSUN, ImageNet) are distinguishable with fewer epochs whereas difficult OOD dataset like CIFAR-100 takes more time. In comparison with the state-of-the-art i.e. convolution \cite{hendrycksUsingSelfSupervisedLearning2019} or contrastive \cite{sehwag2021ssd}, our proposed OODformer converges significantly faster, even with much less batch size. This promising result shows the efficacy of the OODfromer in a real-world scenario and directs to further scope of research of transformer in outlier detection.

\begin{figure}[ht]
\centering
\includegraphics[width=0.45\textwidth]{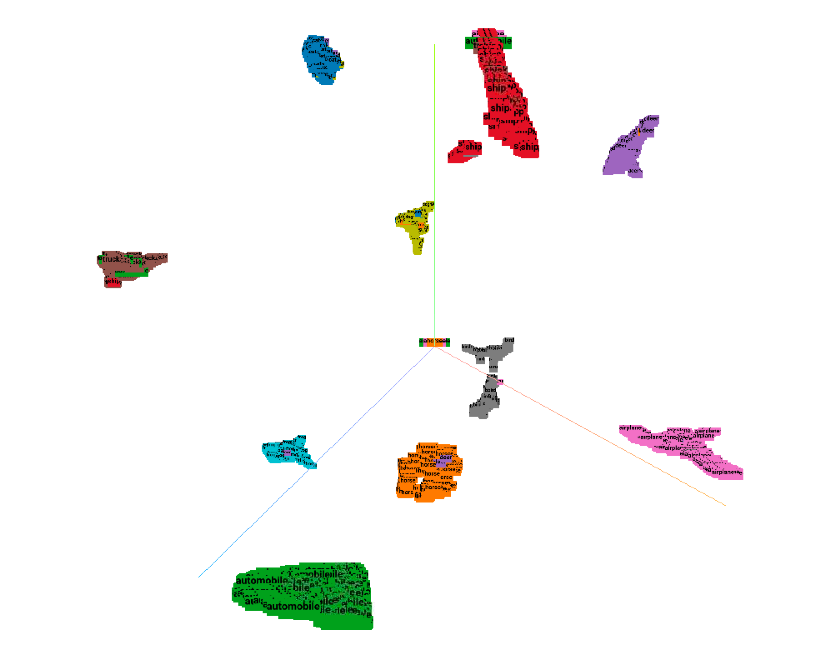}
\includegraphics[width=0.45\textwidth]{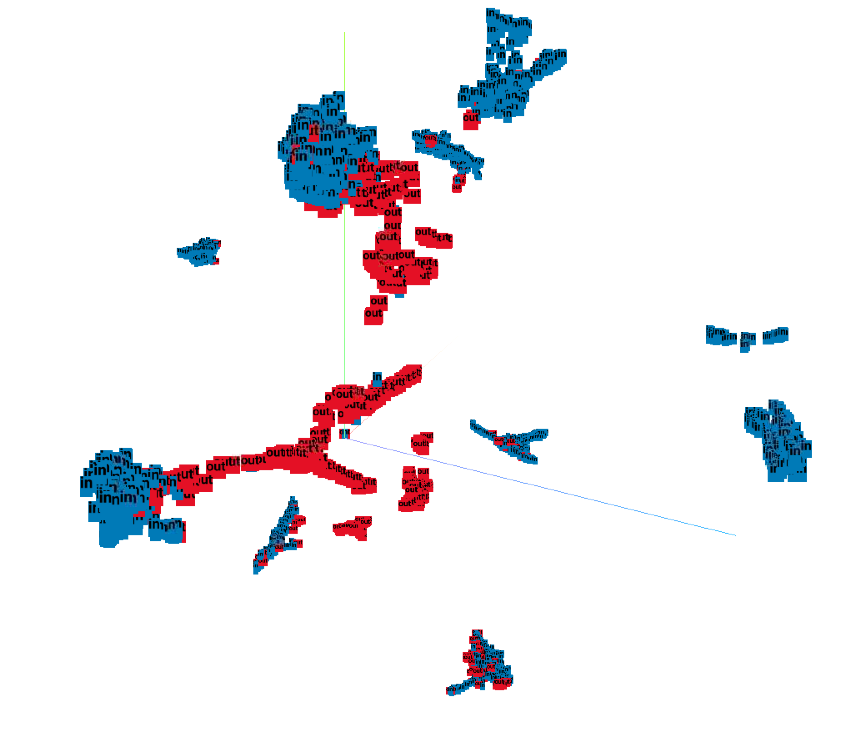}\\
\hspace{0.1cm}(a) \hspace{5cm} (b) 
\caption{OODformer UMAP analysis: a) ID (CIFAR-10) samples and their corresponding cluster, b) ID (blue) and OOD (red) samples shown in UMAP clustering. }
\label{fig:umap}
\end{figure}

\begin{figure}[t!]
\centering
\includegraphics[width=0.45\textwidth]{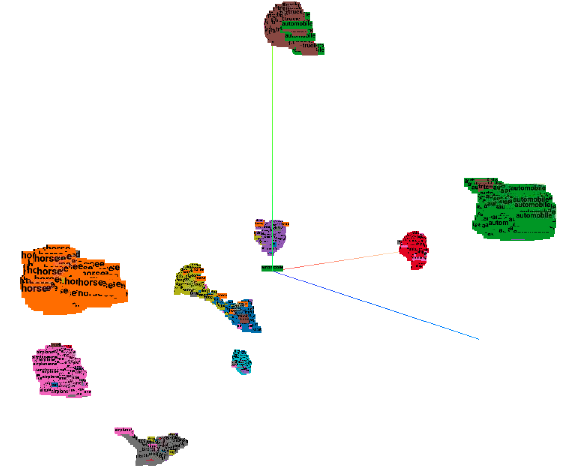}
\includegraphics[width=0.45\textwidth]{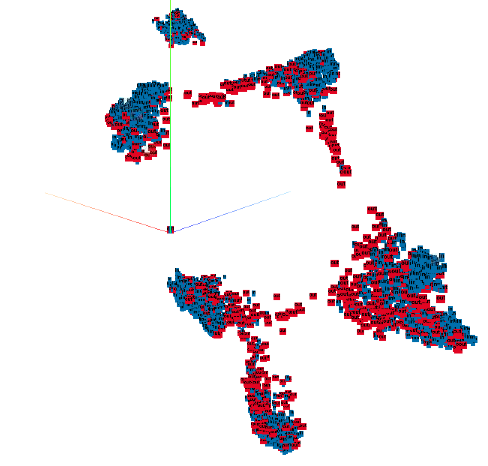}\\
\hspace{0.1cm}(a) \hspace{5cm} (b) 
\caption{ResNet-50 baseline UMAP analysis: a) ID (CIFAR-10) samples and their corresponding cluster, b) ID (blue) and OOD (red) samples shown in UMAP clustering. }
\label{fig:r50umap}
\end{figure}

\textbf{Manifold Analysis :} Fig. \ref{fig:umap}a and \ref{fig:r50umap}a, shows both for OODformer and ResNet-50 baseline, all the classes in CIFAR-10 have formed a compact cluster as shown by their corresponding UMAP. As discussed in Sec. \ref{sec:methods}, we can observe supervise loss helps in the formation of the compact clustering, which can be exploited for class conditioned OOD detection provided there is a separability between ID and OOD data. Figure. \ref{fig:umap}b, shows that for OODformer, OOD samples in the embedding space lie far from any cluster center of an in-distribution sample due to its large distributional shift or lack of object-specific attributes. This variation of distance between an ID and OOD sample is effectively utilized by our distance metric. However, Fig. \ref{fig:r50umap}, suggests that despite being able to form a distinctive cluster for ID samples, our ResNet baseline has failed to maintain a clear separation between an ID and OOD samples. 

This UMAP analysis supports our earlier assumption on results of Table \ref{tab:vitvsres}, in spite of lower or similar accuracy for classification of ID samples, features extracted from transformer have more distinctive separable features for OOD detection.    

\makeatletter
\setlength{\@fptop}{0pt}
\makeatother
\end{document}